\documentclass[runningheads]{llncs}
\usepackage{graphicx}
\usepackage{amsmath,amssymb} 
\usepackage{color}
\usepackage[width=122mm,left=12mm,paperwidth=146mm,height=193mm,top=12mm,paperheight=217mm]{geometry}

\usepackage{bbm}

\usepackage{color,soul}

\usepackage{dsfont}
\newcommand{\real}{\mathbb{R}}

\begin{document}


\pagestyle{headings}
\mainmatter

\title{Adviser Networks:\\
Learning What Question to Ask for Human-In-The-Loop Viewpoint Estimation
} 

\titlerunning{Adviser Networks}

\authorrunning{M. El Banani and J. J. Corso}

\author{Mohamed El Banani and Jason J. Corso}

\institute{University of Michigan\\
Ann Arbor, Michigan, USA\\
\email\{mbanani, jjcorso\} @umich.edu
}

\maketitle

\begin{abstract}



Humans have an unparalleled visual intelligence and can overcome visual ambiguities that machines currently cannot. Recent works have shown that incorporating guidance from humans during inference for monocular viewpoint-estimation can help overcome difficult cases in which the computer-alone would have otherwise failed. These hybrid intelligence approaches are hence gaining traction. However, deciding what question to ask the human at inference time remains an unknown for these problems.
\newline

We address this question by formulating it as an \emph{Adviser Problem}: can we learn a mapping from the input to a specific question to ask the human to maximize the expected positive impact to the overall task? We formulate a solution to the adviser problem for viewpoint estimation using a deep network where the question asks for the location of a keypoint in the input image. We show that by using the Adviser Network’s recommendations, the model and the human outperforms the previous hybrid-intelligence state-of-the-art by {3.7}\%, and the computer-only state-of-the-art by {5.28}\% absolute.


\end{abstract}

\section{Introduction}

Recent years have witnessed great progress on the performance of computer vision algorithms on tasks such as object classification and detection, with state-of-the-art models achieving super-human performance on many of these tasks~\cite{he2015delving,ioffe2015batch}.
In addition, for more challenging tasks, such as viewpoint estimation and fine-grained classification, the community has found that a hybrid intelligence approach---incorporating both human and machine intelligence---outperforms ``computer-only'' approaches~\cite{branson2014ignorant,szeto2017click}.

While hybrid approaches often outperform fully-automated approaches, they do not scale as well;
it is easy to run a model 10x more times, but it can be expensive to ask a human 10x more questions. A clear solution to this problem is to ask the human better questions.

\begin{figure}[t!]
  \begin{center}
  \includegraphics[width=\linewidth]{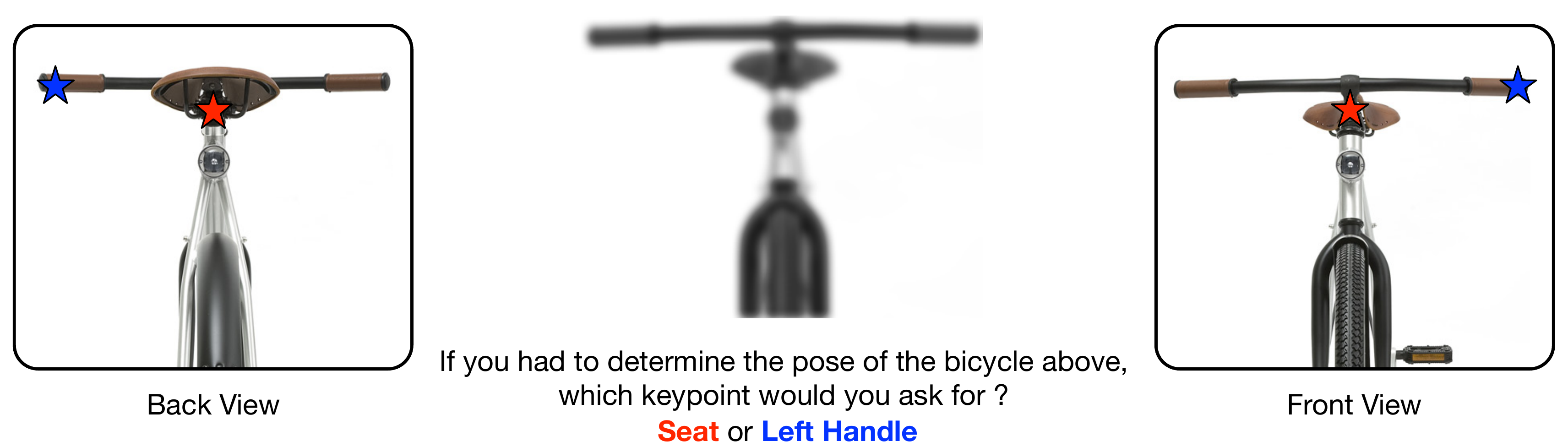}
  \end{center}
     \caption{Some keypoints are more informative than others. 
     			Accurate pose estimation of the bicycle in the center above is difficult due to the blurred out image features. 
                Given access to an oracle, it is clear that one should ask for the location of the left handle rather than the seat to resolve the ambiguity. }
  \label{fig:motivation}
\end{figure}

Consider the task of determining the pose of the bicycle shown at the center of Figure~\ref{fig:motivation}.  Due to its symmetry and truncation, pose estimation algorithms find such examples challenging~\cite{su2015render}. Szeto and Corso~\cite{szeto2017click} have shown that providing such a system with the location and identity of a keypoint on the object allows it to better tackle those examples.  However, not all keypoints are equally informative. While the location of the seat would be the same in both possible poses, the location of the left handlebar allows one to break the symmetry which resolves the ambiguity.  Hence, some queries will allow the hybrid-model to perform better than others. 

In this paper, we ask the question:
how can we determine the best question for a hybrid computer vision model to ask a human?
One approach is to find the question that would maximize information gain with respect to the model's belief~\cite{branson2014ignorant,branson2010visual} or an arbitrary utility function~\cite{russakovsky2015best}.
While those approaches can handle a wide range of inputs,
they require knowledge of the conditional probability functions between the input and/or intermediate variables,
which makes it difficult to handle high-dimensional inputs such as images~\cite{branson2014ignorant}.
They also require the model to have some explicit representation of its belief about the current image.
However, deep learning deals with high-dimensional inputs in an end-to-end manner,
such that the structure to the problem is learned implicitly.
Hence, it is not clear how extendable the previous approaches are to modern deep learning computer vision algorithms.

Ideally, we would want to assume nothing about the hybrid-intelligence model, and deal with it as a black-box that maps inputs to outputs.
In the black-box case, we would have access to nothing but the visual input to the algorithm, the query that was posed to the human, and the algorithm's final output.
In that case, we observe that utility and information gain are correlated with expected task performance, regardless of the nature of that task.
Hence, the problem becomes a classification problem where the input is the visual input to the hybrid-intelligence model and the classes are the queries that the model can pose to the human.

To that end, we propose the Adviser Network; a network
that recommends to its advisee---the hybrid-intelligence model---the question that would result in the highest performance.
We explore Adviser Networks on the task of monocular viewpoint estimation---the problem of identifying the camera's pose (azimuth, pitch and roll angles) with respect to the target object from a single RGB image.
Our choice of task is motivated by multiple reasons.
First, accurate and precise viewpoint estimation is difficult for both humans and machines.
While it is difficult for humans to estimate the exact viewpoint, they can easily guess the overall orientation of an object even if it is symmetrical, occluded, and/or truncated objects.
Computer vision models can precisely estimate the viewpoint of clear asymmetrical objects~\cite{su2015render}, but will struggle with these difficult cases.
Second, the current state-of-the-art performance on a subset of Pascal 3D+~\cite{xiang_wacv14} is achieved by a hybrid intelligence model~\cite{szeto2017click}.
The difficulty and the high performance already achieved on the task make it a challenge for Adviser Networks.

We formulate the problem of choosing the best keypoint for a given image as a classification problem and train an Adviser Network to estimate the expected relative performance for all the possible keypoints for that object. We show that through using the keypoint suggestion from the Adviser Network, the advisee is able to outperform the previous state-of-the-art (which uses a random keypoint) by {3.7}\%, and outperform the computer-only state-of-the-art by {5.28}\%.


In summary, our contributions are as follows:
\begin{itemize}
\itemsep0em
\topsep0em
\item Adviser Networks: Framing the selection of the best question to ask a human as a classification problem,
and proposing a method to train a classification model.
\item Applying Adviser Networks to 3D pose estimation and achieving state-of-the-art performance
on the motor vehicle classes in Pascal 3D+.
\end{itemize}


\section{The Adviser Problem}
\label{sec:adviserProblem}

Consider a hybrid-intelligence algorithm: one that can compute the solution to a problem and, while doing so, incorporate guidance from an oracle (a human) during computation~\cite{merritt_kurator_2017}. What guidance is provided by the human, however, is unclear.  Some guidance may lead to performance improvements while others may have little or even negative impact.

The role of the Adviser is to seek the best guidance possible from the human,
which is the guidance that will ultimately lead to the largest performance improvement of the hybrid-intelligence algorithm.
We hence call that algorithm the \textit{advisee}.
However, the Adviser must make a choice about what guidance to seek from the input data only;
it cannot have the advisee compute all possible outcomes and then select among them post-hoc as this would require too many queries to the human and too many evaluations of the advisee model. Let us now make this idea more concrete.

\paragraph{Problem Statement}
%
%
Let algorithm $f$ be a viewpoint estimation algorithm that maps an image, $x$, and human-guidance to an output, $y$.
The human-guidance is the response to a query, $q$, from a finite set of queries, $Q$; in our notation, we use $q$ both for the query and the human's response for simplicity.
The output of $f$ is evaluated by an function, $R$, that maps the algorithm's output, $y$, along with the ground-truth value, $z$, to a scalar value that captures its performance.

The goal of the Adviser is to learn a mapping from $x$ to $q \in Q$ to maximize the evaluation $R(f(x,q),z)$:
\begin{align}
\label{eq:optimizationFn}
p_Q(x) \doteq q^* = \arg \max_{q\in Q} \mathds{E}\left[ R(f(x,q), z) \right]
\end{align}
where $p_Q(x)$ represents the Adviser network, while $f(x,q)$ represents the advisee network.

\subsection*{Concrete Advisee Example: Human-guided Viewpoint Estimation}
We focus on the task of human-guided monocular viewpoint estimation. Here, we are given a single image as input, $x \in \real^{h\times w \times 3}$,
and assume it contains one instance of an object from set $\mathcal{C}$, such as a car or a bicycle.
We are also provided with the location of a single keypoint from the set $K_c$; 
the set $K_c$ for an instance of $c \in \mathcal{C}$ contains all the possible keypoints for object $c$.  In the case of the car category, the keypoint set may contain \textit{front-left-bumper} and \textit{rear-right-wheel} among others. For simplicity, we overload $q$ to be the answer to such a query and hence, $q$ is a tuple $(k,u,v)$ where $k\in K_c$ is the identity of the keypoint and $(u,v)$ is its pixel location in the input image $x$.

The challenge of human-guided viewpoint estimation is to integrate those two inputs to estimate the three-degrees-of-freedom viewpoint of the camera with respect to that object in the image, which we call \textit{pose} for the remainder of the paper: the azimuth $\theta_1$, pitch $\theta_2$, and roll $\theta_3$ angles.  So, $y = \left(\theta_1, \theta_2, \theta_3\right)$.  The evaluation function $R(\cdot)$ is the geodesic distance between the predicted pose and the ground-truth pose:
\begin{align}
R(y,z) = \frac{\lVert\log(\mathbf{R}_y^{\textsf{T}} \mathbf{R}_z)\rVert_F}{\sqrt{2}}
\label{eq:geodesic}
\end{align}
where $\mathbf{R}_y$ is the rotation matrix induced by predicted pose $y$ and $\mathbf{R}_z$ is the rotation matrix induced by ground-truth pose $z$ and $||\cdot||_F$ is the Frobenius norm.

For our experiments, we use the Click Here CNN~\cite{szeto2017click} model which currently has the state-of-the-art performance on the vehicle classes of Pascal 3D+~\cite{xiang_wacv14}.  A Click Here CNN leverages human-guidance by learning to use it to spatially attend to more important areas of the convolved image features.  We discuss this example in full detail in Section \ref{sec:viewpoint}.

\paragraph{Advisers in Practice: Adviser Networks}
When formulating an instance of the Adviser problem in practice, we realize that seeking an explicit $p_Q(x)$ is ambitious: it may not even be directly evaluated as the probability density of the evaluation is not known and the full evaluation would, again, require too much human input.  We hence elect to learn a data-driven deep network to \textit{approximate} it (not in a rigorous sense).

A key question that arises in specifying such an Adviser Network is whether we want to estimate the absolute expected error (evaluating $R$) associated with each possible query $q \in Q$ for a given input $x$, or if we only care about the relative performance of each $q$.  If we setup the Adviser network to learn to predict the relative, rather than absolute, performance of each possible query $q$ then our predictions become robust to the actual difficulty of a given instance.

For example, consider our viewpoint estimation case for the object category of bicycle.  Now, we may have a blurry image and a well-focused image.  If we seek a prediction of the absolute performance of a particular keypoint query, then the network will need to focus on the specific properties of the input image yielding a harder problem (considering the blurry versus well-resolved inputs) whereas, if we seek relative performance, then the network need not focus on these specifics and rather seek a more general representation over the image.  In this case, for example, the network learning relative performance may learn to prefer a query for the \textit{left handle} when trying to disambiguate front and back views of a bicycle.

In this work, we implement Eq.~\ref{eq:optimizationFn} as a classification problem on $q$; where the Adviser Network is tasked with predicting a probability distribution over the expected relative impact of specific keypoints.
We support our argument for the use of classification in Section~\ref{sec:regression} by showing that learning to predict the relative performance allows the Adviser Network to propose better keypoints than learning to estimate the absolute performance.  We discuss the application of Adviser Networks to viewpoint estimation in more detail below.  In section~\ref{sec:regression}, we show that learning to predict such relative performance allows Adviser networks to achieve a better performance than by trying to estimate the absolute error.


\section{Viewpoint Estimation with an Adviser}
\label{sec:viewpoint}
\begin{figure*}[t]
  \begin{center}
  \includegraphics[width=0.9\linewidth]{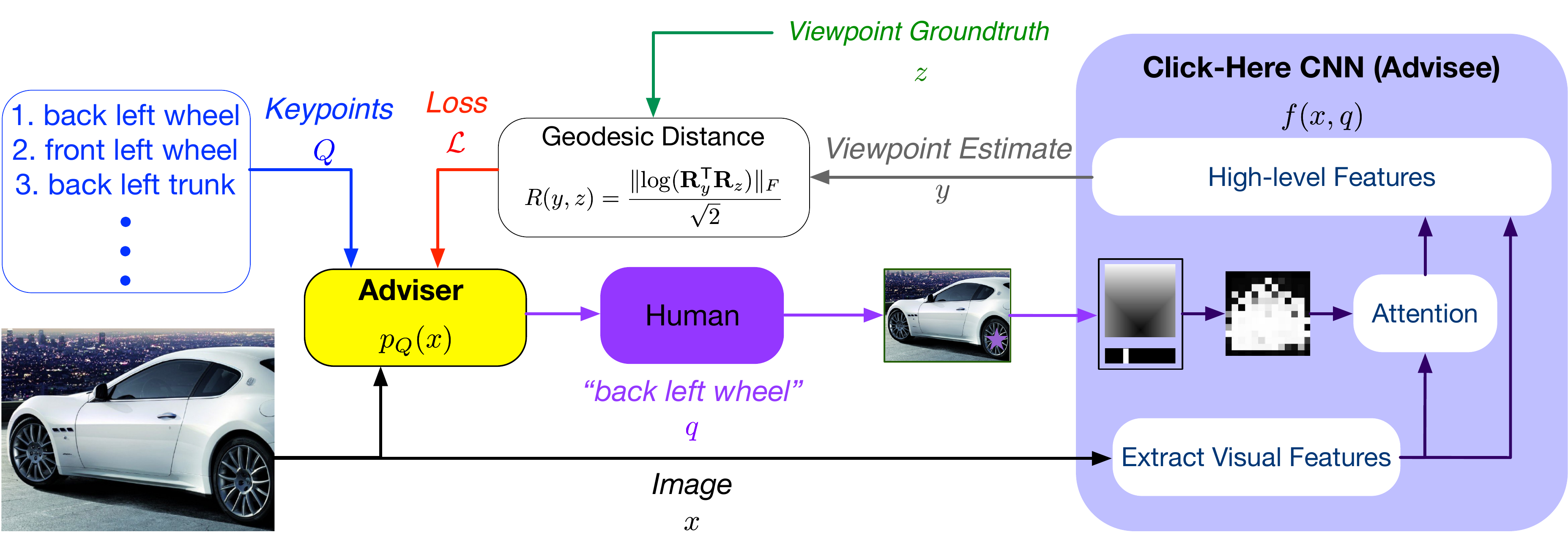}
  \end{center}
     \caption{Instantiation of Adviser model for the monocular viewpoint estimation. We use key-point location as human input that helps Click Here CNN to estimate viewpoint prediction better than without human input. The Adviser Network selects the key-point class to query the human.}
  \label{fig:adviserNetwork}
\end{figure*}

We apply an Adviser Network to the task of viewpoint estimation, as shown in Figure~\ref{fig:adviserNetwork}.  The goal of the Adviser Network is to choose the keypoint that would result in the most accurate viewpoint estimate, or pose, by the network. As mentioned in Section~\ref{sec:adviserProblem}, an adviser problem is defined by the advisee algorithm, the set of queries that it can pose to a human, and an evaluation function.

In this work, we use Click Here CNN~\cite{szeto2017click} as our advisee network.  The Click Here CNN network has two streams: an image stream composed of the first seven layers of AlexNet~\cite{krizhevsky2012imagenet} and a keypoint-weighted stream which encodes the human guidance as a weighting mask that is used to modulate the fourth convolutional layer in the image stream.  The output of the two streams are concatenated and then fully connected layers are used to ultimately estimate the pose.  See \cite{szeto2017click} for a full description of Click Here CNN.

This choice of the advisee constrains the set of queries to the keypoint classes for the three motor-vehicle classes in Pascal 3D+~\cite{xiang_wacv14}. We define our evaluation function as the geodesic distance function (Equation~\ref{eq:geodesic}) due to its common use in the viewpoint estimation literature~\cite{su2015render,tulsiani2015viewpoints}. It should be noted that our choice of evaluation function is not limited to purely performance metrics; it is possible to define an evaluation function that incorporates some measure of utility~\cite{russakovsky2015best} or query difficulty~\cite{branson2014ignorant}.

We use a variant of AlexNet~\cite{krizhevsky2012imagenet} as the architecture of our Adviser network, with the final layer modified to output 34 classes, i.e. the number of keypoint classes for the 3 motor vehicle classes in Pascal 3D+.  We initialize the Adviser Network with the weights from the image stream of Click Here CNN in order to take advantage of the learned representations.  Since our Adviser network is predicting the keypoint performance for three different object classes at the same time, we adopt the approach of Tulsiani and Malik~\cite{tulsiani2015viewpoints} and use a loss layer that applies the loss selectively to the keypoints belonging to the object class of the training instance. Hence, we are able to jointly train the adviser network on multiple object classes.

In order to classify images to keypoints, we need the convert the output of the evaluation function to a probability for each keypoint that captures how informative we expect the keypoint to be. We calculate a structured label, $y_x$, for image $x$ by using a weighted softmax:
\begin{equation}
y_{x} =
\begin{cases}
\frac{\exp(-R(f(x,q_i), z)/T)}{\sum_{j \in KP(I)}{\exp(-R(f(x,q_j), z)/T)}}	& \text{if } i \in KP(I) \\
0 & \text{otherwise, }
\end{cases}
\label{eq:performanceToProb}
\end{equation}
where $R(f(x,q_i), z)$ is the geodesic error (Eq.~\ref{eq:geodesic}) associated with the i-th keypoint on the image when it is used by Click Here CNN,
$KP(I)$ is the set of keypoints that are visible on image $I$,
and $T$ is a temperature to scale the error~\cite{hinton2015distilling}.
To illustrate an example, in the extreme case where two keypoints perform well,
while all the others perform poorly,
the probability distribution across keypoints would be near 50\% for the two good keypoints,
near 0 for the bad keypoints, and exactly 0 for all keypoints that are not visible on the image.

We use a temperature parameter in a similar spirit to Hinton \textit{et al.}~\cite{hinton2015distilling}.
One difference is that they restrict their use to temperature values larger than 1 to soften the labels, while we use a wider range of values to modulate performance differences between keypoint errors. Different tasks will map their performance to different ranges, \textit{e.g.} geodesic errors reported in degrees will be larger that those reported in radians, despite being exactly the same. Hence a scaling factor is needed to correct for that. In our experiments, we have found that the model had equivalent performance for temperature values between 0.1 and 10, and we used a temperature value of 10 for all our experiments.  

\section{Experimental Setup}
\begin{figure}[t!]
	  \includegraphics[width=\linewidth]{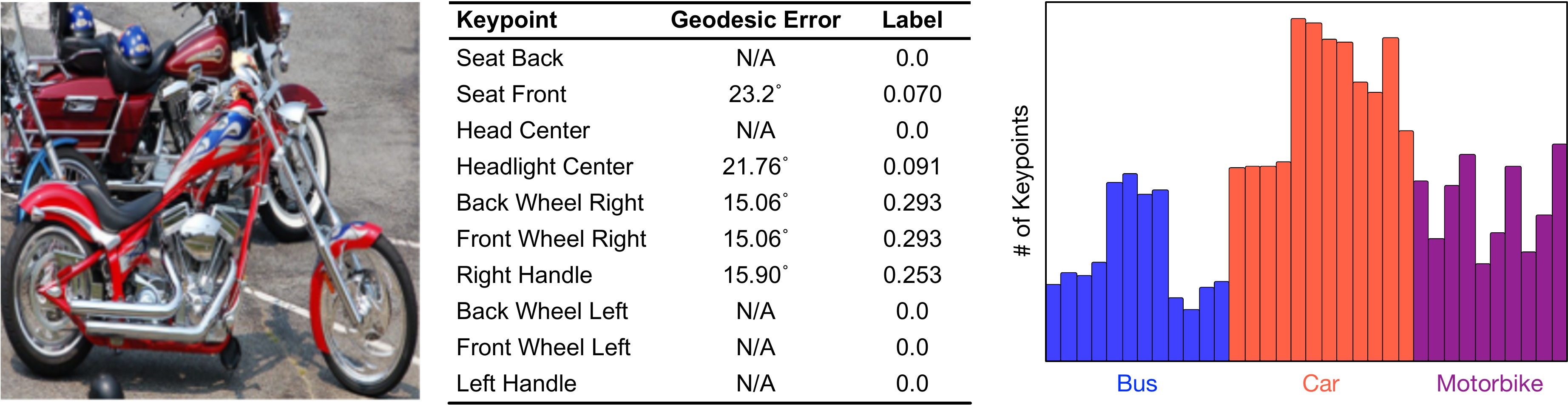}
     \caption{Advisee example for Geodesic Error and Label for each keypoint class. The label was generated using a temperature value of 0.01. The histogram on the right show the keypoint distribution for each of the classes}
  \label{fig:adviseeExample}
\end{figure}

\paragraph{Dataset}
We evaluate Adviser Networks on the motor vehicle subset of Pascal 3D+~\cite{xiang_wacv14};
buses, cars, and motorbikes.
We use the train/test split provided in the dataset.
The motor vehicle subset includes 1554 training examples and 1552 testing examples.
Dataset instances in Pascal 3D+ are annotated for difficulty, truncation, and occlusion.
Previous work has often restricted their evaluation to non-truncated and non-occluded instances~\cite{su2015render,tulsiani2015viewpoints}.
In this work, we use the full test set to demonstrate the robustness of our approach.

The advisee dataset is generated using the procedure outlined in Section~\ref{sec:viewpoint}.
We use the Click Here CNN as the advisee network.
The performance for each keypoint is characterized by the distance between the estimated and ground-truth viewpoints.
An example of an instance of the advisee set can be seen in Figure~\ref{fig:adviseeExample}.
We follow the same approach used by previous work on viewpoint
estimation~\cite{szeto2017click,su2015render,tulsiani2015viewpoints}
using the geodesic distance as our evaluation metric and reporting the median geodesic error and accuracy; where accuracy is calculated as the fraction of instances whose geodesic error is lower than $\pi/6$.

Since the Click Here CNN model is fine-tuned on Pascal 3D+, the network performs significantly better on the training data than on the test data. This results in a lower variance in the keypoint performance on the training data, which greatly affects the representations learned by the Adviser Network; the Adviser Network cannot learn which keypoints are more informative if the training accuracy is 100\% as there will be no performance difference between the different keypoints. 
We tackle this problem in two ways, we generate the training set using a less accurate version of Click Here CNN---one that is not fully fine-tuned on Pascal 3D+ and that has a training accuracy comparable to the final testing accuracy. This allows us to train a model that can be evaluated on the full Pascal 3D+ test set. We also define another dataset, Pascal-Small, which is generated by splitting the Pascal Test set into 70\% subset used for training and 30\% subset using for testing. Pascal-Small is intended to show that our approach can be implemented using only a single black-box model, without access to intermediate model weights or the training data on which the model was trained.    

\paragraph{Training and Testing}
We use the same training regime for all our experiments unless otherwise stated.
We fine-tune the adviser network by training all layers using randomly sampled batches of size 256.
The images are preprocessed by normalizing the RGB channels with the mean values from ImageNet~\cite{russakovsky2015imagenet} and cropping them using ground-truth bounding boxes.
We train each model for 100 epochs and report the final test accuracy. We choose not to use early stopping since all models converge to their final performance.
We evaluate models by considering the performance of Click Here CNN using the top applicable keypoint prediction from the Adviser network. A keypoint is applicable if it visible on the image. For regression, we evaluate the model in a similar fashion by ranking the errors by their value and taking the one that is expected to produce the minimum error. All the models are trained using SGD optimizer~\cite{kingma2015adam} with a learning rate of $10^{-2}$,
a momentum of 0.9, a weight decay of $5\cdot 10^{-4}$, and a learning decay rate of 0.95 every 5 epochs.
We did not perform any fine-tuning of hyper-parameters.
All the code is implemented in PyTorch~\cite{paszkepytorch} and will be released upon publication.

While it is common to use cross entropy as a classification loss, we found that it did not work well; models trained using cross-entropy did uniformly worse than models trained using mean-square error, and overall performance was poor. This finding has been previously reported by Hinton \textit{et al.}, where they reported that a mean-square error loss works better than binary cross entropy in matching the soft probabilities of a model~\cite{hinton2015distilling}. 

\section{Experimental Results}

We evaluate the Adviser network by evaluating the performance of Click Here CNN when it uses the recommendations of the Adviser Network.
We note that Click Here CNN was evaluated on an augmented dataset where instances were image-keypoint pairs. Hence, to compare against Click Here, one needs to reduce all the instances belonging to the same image onto a single instance. Therefore, we compare against seven baselines:
\begin{itemize}
\itemsep0em
\topsep0em
\item \textbf{Render For CNN}: 		The performance of the computer-vision only state of the art model~\cite{su2015render}.
\item \textbf{Render For CNN-FT}: 	The performance of the computer-vision only state of the art model after fine-tuning it on the vehicle classes.
\item \textbf{Lower-bound}: 		maximum geodesic error across keypoints for each image. This can be thought of as an adversarial Oracle.
\item \textbf{Upper-bound}: 		minimum geodesic error across keypoints for each image. This can be thought of as a friendly Oracle.
\item \textbf{Click Here CNN}: 		the mean geodesic error for all the keypoints in the image. This is the expected performance of Click Here CNN~\cite{szeto2017click}.
\item \textbf{Frequency Prior}: 	pick the keypoint that has appeared the most in the test set.
\item \textbf{Performance Prior}: 	pick the keypoint that has performed the best on the test set.
\end{itemize}

We note that both priors were constructed using Click Here's performance on the test set, hence they are very strong priors. When evaluating the priors, we pick the highest ranking keypoint according the prior that exists on the image.
The performance values calculated for Render For CNN and Click Here CNN are different to those reported by the authors in the respective papers, as the authors of those papers evaluate on variants of Pascal 3D+ and not the actual dataset. 
Su \textit{et al.}~\cite{su2015render} focus on evaluating the viability of training a module using synthetic data only and evaluate that model on the ``easy" susbet of Pascal 3D+ which excludes any objects that are truncated, occluded, or not clearly visible~\cite{su2015render}. Szeto and Corso~\cite{szeto2017click} evaluate the performance on a variant of Pascal 3D+ where each data instance is an image-keypoint tuple. Hence, all their reported results are for that augmented dataset. To generate their results, we used the trained model weights that those authors have made publicly available on github. 


\begin{table}[b]
\begin{center}
   \caption{Click Here CNN performance with and without Adviser on the full Pascal 3D+ Vehicles test set}
\begin{tabular}{  l | c c c | c || c c c | c }
    \multicolumn{1}{}{}
    & \multicolumn{4}{c}{ $Accuracy_{\pi/6}$ }
    & \multicolumn{4}{c}{ $Median$ $Geodesic$ $Error$}\\
    \hline
    									& Bus   & Car   & M.bike 	& Mean  & Bus   & Car   & M.bike  	& Mean  \\
    \hline
    \hline
    Render For CNN
    & 89.32 & 78.39 & 76.99  	& 81.57 &  5.21 & 8.29  & 15.00   	& 9.50	\\
    Render For CNN-FT 					& 88.26 & 80.00 & 83.48  	& 83.91 &  3.61 & 6.83  & 12.22    	& 7.55	\\
    \hline
    Lower-bound         				& 88.61	& 82.37	& 79.65 	& 83.54 & 3.81  & 6.63  & 13.93 	& 8.12	\\
    Click Here CNN
& 90.04	& 85.59	& 80.83 	& 85.49 & 3.54  & 6.17  & 13.41 	& 7.71  \\
    Frequency Prior     				& 93.59	& 87.63				& 82.01 	& 87.74 & 3.51  & 5.78  & 13.27 	& 7.52	\\
    Performance Prior   				& \textbf{93.95}	& 87.96				& 83.78 	& 88.56 & 3.54  & 5.77  & 12.93 	& 7.41  \\
	Adviser (Ours)						& \textbf{93.95}  	& \textbf{89.25}  	& \textbf{84.37}  & \textbf{89.19}  & \textbf{3.48}  & \textbf{5.75}  & \textbf{12.89}  & \textbf{7.37}		\\
    \hline
    Upper-bound         				& 95.02	& 92.47	& 87.32 	& 91.60 & 3.00  & 5.32  & 11.76 	& 6.69	\\
    \hline
    \end{tabular}
	\end{center}
\label{table:pascalFullResults}
\end{table}

\subsection{Pascal3D+ Motor Vehicles}

The results on the full Pascal test set are presented in Table~\ref{table:pascalFullResults}. We show that using the Adviser Network outperforms the expected performance of Click Here across all metrics, with a mean improvement of {3.7}\%. This finding supports our claim that an Adviser network can improve human-in-the-loop visual inference. Despite the performance prior exceeding the Adviser performance on some metrics, the Adviser network still has a better mean performance for both accuracy and geodesic error.


\subsection{Pascal-Small}

In order to provide a more rigorous empirical validation of our model, we train it on a random subset of a test set and test it on the remaining subset. To account for the randomness of the split, we ran 6 different experiments: 3 experiments using the training set and 3 experiments using the test set. In our experiments we use a 70:30 train-test split. The aggregated results, mean and standard deviation, of those experiments are presented in Table~\ref{table:pascalSmallResults}. The results show that Adviser method outperforms Click-Here performance and both priors. This provides some support that the method provides an improvement over all methods when both training and test data have a similar distribution. 

It should be noted that the generated datasets are extremely small; the training data has less than 1000 instances. Under such conditions, fully-supervised methods have a huge disadvantage against a non-parametric baseline like a performance prior.

\begin{table}[t]
\begin{center}
   \caption{Aggregated Click Here CNN performance with and without Adviser on different Pascal Small subsets}
    \begin{tabular}{  l | c c c | c }
    \hline
    									& Bus   & Car   & M.bike  	& Mean  \\
    \hline
    \hline
    Lower-bound         				&	89.28 $\pm$ 3.01   &   87.36 $\pm$ 4.21  &    81.70 $\pm$ 3.43  &    86.11 $\pm$ 3.31 \\
    Click Here CNN~\cite{szeto2017click}&	92.19 $\pm$ 3.74   &   90.64 $\pm$ 4.12  &    83.18 $\pm$ 3.99  &    88.67 $\pm$ 3.86 \\
    Frequency Prior     				&	95.69 $\pm$ 3.15   &   91.78 $\pm$ 3.83  &    83.97 $\pm$ 3.47  &    90.48 $\pm$ 3.24 \\
    Performance Prior   				&	95.56 $\pm$ 2.90   &   92.03 $\pm$ 3.36  &    84.95 $\pm$ 2.20  &    90.85 $\pm$ 2.56 \\
	Adviser (Ours)						&	\textbf{95.93 $\pm$ 3.00}   &   \textbf{92.68 $\pm$ 3.33}  &    \textbf{85.05 $\pm$ 2.26}  &   \textbf{91.22 $\pm$ 2.64} \\
    \hline
    Upper-bound         				& 97.13 $\pm$ 2.55   &   95.85 $\pm$ 3.08  &    86.23 $\pm$ 1.97  &    93.07 $\pm$ 2.16 \\
    \hline
    \end{tabular}
	\end{center}
\label{table:pascalSmallResults}
\end{table}

\subsection{Keypoint Classification vs Error Regression}
\label{sec:regression}

The Adviser Network could learn the importance of different keypoints by either learning to predict their relative performance via classification or through learning to regress the error associated with each keypoint. It should be emphasized that the discrimination between classification and regression does not depend on the loss being used, but rather on the how the labels are represented.
We test our claim that learning the relative performance allows the model to achieve a higher performance by comparing the classification performance against regression. Since the geodesic error can be calculated in radians or degrees, we train models to regress both errors. 

\begin{table}[h]
    \begin{center}
       \caption{Comparison between the performance of regression and classification loss for Adviser Networks}
        \begin{tabular}{  l | c c c | c || c c c | c  }
\multicolumn{1}{}{}
    & \multicolumn{4}{c}{ $Accuracy_{\pi/6}$ }
    & \multicolumn{4}{c}{ $Median$ $Geodesic$ $Error$}\\
    \hline
     							& Bus   & Car   & Motorbike	& Mean  & Bus   & Car   & Motorbike & Mean  \\
\hline
        {Classification}		& \textbf{93.95}  	& \textbf{89.25}  	& 84.37  			& \textbf{89.19}  	& \textbf{3.48}  & \textbf{5.75} & 12.89  			& \textbf{7.37} 	\\ 
        {Regression (degrees)}	& 91.1  			& 87.42  			& \textbf{84.66}  	& 87.73  			& 3.61  		 & 5.8  		 & 12.94  			& 7.45		\\ 
        {Regression (radians)}	& 91.1  			& 87.74  			& \textbf{84.66}  	& 87.83  			& 3.61  		 & 5.81  		 & \textbf{12.69}  	& \textbf{7.37}	\\ 
\hline
        \end{tabular}
    \end{center}
\label{fig:regressionComparisonAll}
\end{table}

As shown in Table~\ref{fig:regressionComparisonAll}, classification outperforms regression for all the classes except for the motorbike class.
While it is not clear why this happens, one reason could be that there is more variance in the geodesic errors for the Motorbike class in both the training set and test set. 
This increased variance could allow the model to learn more by regressing the error, rather than just classifying the best keypoint(s).

\section{Related Work}

There has been a lot of recent interest in using human-in-the-loop for computer vision applications.  While the majority of this work has been done to curate better datasets \cite{russakovsky2015best,kovashka2011actively}, there has been some of research on using human-guidance to tackle some of the more challenging visual tasks such as viewpoint estimation~\cite{szeto2017click} and fine-grained classification~\cite{branson2014ignorant,branson2010visual}.
We focus our comparison against methods that were focused on how to best leverage human guidance within the scope of human-in-the-loop computer vision.
One approach is to find the question that maximizes utility \cite{russakovsky2015best} or information gain \cite{branson2014ignorant}. While those approaches have a strong theoretical basis and can provide remarkable performance, they require one to have an accurate model of the task to calculate the expected utility or information gain.  The generation of such a model can be difficult for highly dimensional inputs~\cite{branson2014ignorant}.
The creation of the model also requires for the task to be clearly understood to assign the correct utilities.
Another approach would be to provide the human with all possible queries and ask them to answer any of them. One could possible adopt this approach if they believe that inputs are equally informative.  However, previous work has shown that this is not always the case \cite{branson2014ignorant,szeto2017click}.  One could also assume that humans will answer the query that is most discriminative.  While Deng \textit{et al.} have shown that this can provide boosts in performance for attribute-based classification~\cite{deng2013fine}, Linsley \textit{et al.} have shown that humans do not focus on the same features as deep convolutional neural networks~\cite{linsley2017visual}. Hence relying on human bias might not result in the optimal performance.

Our work is also related to literature that attempts to learn representations from one network to another.
Bucila \textit{et al.}~\cite{bucilua2006model} show that one can compress the representations learned by a model in smaller model, while Hinton \textit{et al.}~\cite{hinton2015distilling} show that one can transfer the representations learned by a large network to a smaller one. 
Romero \textit{et al.}~\cite{romero2014fitnets} explore the idea of using a network to teach a learned to another network. 
While our approach is similar to this line of research in terms of using two networks to help each other, we differ in a very important way; 
while both their models are learning to perform the same task,
the Adviser Network is trying to learn the meta task of finding the best keypoint instead of replicating the task of its advisee.
In that way, our work is closer to the literature on generative adversarial networks~\cite{goodfellow2014generative}, with the main difference being that our networks are not competing against each other, nor are they co-trained. 

Previous work that greatly shares our overall goal is work related to mixed expert models~\cite{jacobs1991adaptive}.
There a gating network learns to assign different tasks to different networks. A crucial difference that differentiates our work is that the gating network is trained jointly with the expert models, while we assume a black-box model that has already pretrained.

Work in the area of active learning~\cite{settles2011theories} has been concerned with choosing the best instance to train a model. While we share their goal of finding the next data point to choose, their work is focused on finding the best instance to train a model, while our focus is on choosing the best keypoint at inference time. Hence, their methods do not apply to our problem.

Finally, there has been research on improving human-computer interaction and human-robot interaction through asking better questions in tasks such as navigation~\cite{cai2016asking}, task learning~\cite{cakmak2012designing}, and question-answering~\cite{buck2017ask}. While this work shares our goal of learning to best leverage the human input, they differ in problem scope, task representation, and assumed knowledge of the task. 
Cai and Mostofi train a convolutional neural network to predict how easy it is for a human to detect an object in a specific image~\cite{cai2016asking}. The trained network is then used by a robot as it navigates around the environment to determine when it would be useful to query a remote operator about its current surroundings. Their method differs from us in that they are trying to predict human performance, rather than the robots performance on the task. 
Cakmak and Thomaz train a robot to choose which questions to ask to improve its task learning capability~\cite{cakmak2012designing}. While their goal is similar, their focus is on the interaction aspect of the problem; the type of questions humans typically ask and how humans perceive the question-asking robot. Buck \textit{et al.} explore a method to reformulate questions asked by humans to improve a black-box question answering system~\cite{buck2017ask}. While they share our assumption that the \textit{advisee} is a black-box, but they also assume that the question being asked is fixed; their task is to re-represent the question to improve the advisee's performance. This differs from the problem we are tackling which is focused on posing a question to the human. Hence, the overall goal of their work assumes an opposite information flow than ours.


\section{Conclusion and Future Work}

In this work, we define the Adviser problem as the problem of finding the query that would best improve the performance of a hybrid-intelligence system. We formulate a solution to the adviser problem for human-guided viewpoint estimation using a deep convolutional neural network to find the best keypoint to ask of the human. We show that by using the keypoint guidance from the Adviser Network and the human, the model is able to outperform the previous hybrid-intelligence state-of-the-art by 3.7\%, and outperform the computer-only state-of-the-art by 5.28\% absolute.

While we assume that human choice of keypoints is random, this might not be the case. Deng \textit{et al.} have shown that training using human chosen patches can improve classification performance~\cite{deng2013fine}. However, Linsley \textit{et al.} have shown that state-of-the-arts models pay attention to different parts of the image than humans. Hence, it would be beneficial to compare unconstrained human choices to the Adviser Networks output.

While our focus here has been on using Click Here CNN for viewpoint estimation, our approach is easily extensible to other hybrid intelligence tasks due to the generality of our approach. It should be noted that the adviser problem defined in Section~\ref{sec:adviserProblem} is extensible to any human-in-the-loop problem, and is especially well suited for problems where the task is not sufficiently well-modeled and the inputs are high-dimensional.  While such extensions are beyond the scope of this paper, we plan on extending this work to other tasks such as fine-grained classification and video dense-captioning, as well as integrating more complex modalities for human guidance such as language.

\parskip 8pt
\noindent
\textbf{Acknowledgments}
We thank Madan Ravi Ganesh, Shurjo Banerjee, Vikas Dhiman, Sajan Patel, and Ryan Szeto for their helpful discussions.
This work has been partially supported by DARPA W32P4Q-15-C-0070 (subcontract from SoarTech) and funds from the University of Michigan Mobility Transformation Center. 
Jason Corso is a consultant for Soartech, who also partially sponsored this work under contract 10322.01 with prime contract W31P4Q-16-C-0091.

\bibliographystyle{splncs}
\bibliography{egbib}

\begin{thebibliography}{10}

\bibitem{he2015delving}
He, K., Zhang, X., Ren, S., Sun, J.:
\newblock Delving deep into rectifiers: Surpassing human-level performance on
  imagenet classification.
\newblock In: Proceedings of the IEEE international conference on computer
  vision. (2015)  1026--1034

\bibitem{ioffe2015batch}
Ioffe, S., Szegedy, C.:
\newblock Batch normalization: Accelerating deep network training by reducing
  internal covariate shift.
\newblock In: International Conference on Machine Learning. (2015)  448--456

\bibitem{branson2014ignorant}
Branson, S., Van~Horn, G., Wah, C., Perona, P., Belongie, S.:
\newblock The ignorant led by the blind: A hybrid human--machine vision system
  for fine-grained categorization.
\newblock International Journal of Computer Vision \textbf{108}(1-2) (2014)
  3--29

\bibitem{szeto2017click}
Szeto, R., Corso, J.J.:
\newblock Click here: Human-localized keypoints as guidance for viewpoint
  estimation.
\newblock In: Proceedings of IEEE International Conference on Computer Vision.
  (Oct 2017)

\bibitem{su2015render}
Su, H., Qi, C.R., Li, Y., Guibas, L.J.:
\newblock Render for cnn: Viewpoint estimation in images using cnns trained
  with rendered 3d model views.
\newblock In: Proceedings of the IEEE International Conference on Computer
  Vision. (2015)  2686--2694

\bibitem{branson2010visual}
Branson, S., Wah, C., Schroff, F., Babenko, B., Welinder, P., Perona, P.,
  Belongie, S.:
\newblock Visual recognition with humans in the loop.
\newblock In: Proceedings of the European Conference on Computer Vision,
  Springer (2010)  438--451

\bibitem{russakovsky2015best}
Russakovsky, O., Li, L.J., Fei-Fei, L.:
\newblock Best of both worlds: human-machine collaboration for object
  annotation.
\newblock In: Proceedings of the IEEE Conference on Computer Vision and Pattern
  Recognition. (2015)  2121--2131

\bibitem{xiang_wacv14}
Xiang, Y., Mottaghi, R., Savarese, S.:
\newblock Beyond pascal: A benchmark for 3d object detection in the wild.
\newblock In: Proceedings of the IEEE Winter Conference on Applications of
  Computer Vision (WACV). (2014)

\bibitem{merritt_kurator_2017}
Merritt, D., Jones, J., Ackerman, M.S., Lasecki, W.S.:
\newblock Kurator: Using the crowd to help families with personal curation
  tasks.
\newblock In: Proceedings of the 2017 ACM Conference on Computer Supported
  Cooperative Work and Social Computing. CSCW '17, New York, NY, USA, ACM
  (2017)  1835--1849

\bibitem{krizhevsky2012imagenet}
Krizhevsky, A., Sutskever, I., Hinton, G.E.:
\newblock Imagenet classification with deep convolutional neural networks.
\newblock In: Advances in Neural Information Processing Systems. (2012)
  1097--1105

\bibitem{tulsiani2015viewpoints}
Tulsiani, S., Malik, J.:
\newblock Viewpoints and keypoints.
\newblock In: Proceedings of the IEEE Conference on Computer Vision and Pattern
  Recognition. (2015)  1510--1519

\bibitem{hinton2015distilling}
Hinton, G., Vinyals, O., Dean, J.:
\newblock Distilling the knowledge in a neural network.
\newblock arXiv preprint arXiv:1503.02531 (2015)

\bibitem{russakovsky2015imagenet}
Russakovsky, O., Deng, J., Su, H., Krause, J., Satheesh, S., Ma, S., Huang, Z.,
  Karpathy, A., Khosla, A., Bernstein, M.,  et~al.:
\newblock Imagenet large scale visual recognition challenge.
\newblock International Journal of Computer Vision \textbf{115}(3) (2015)
  211--252

\bibitem{kingma2015adam}
Kingma, D., Ba, J.:
\newblock Adam: A method for stochastic optimization.
\newblock (2015)

\bibitem{paszkepytorch}
Paszke, A., Chintala, S., Collobert, R., Kavukcuoglu, K., Farabet, C., Bengio,
  S., Melvin, I., Weston, J., Mariethoz, J.:
\newblock Pytorch: Tensors and dynamic neural networks in python with strong
  gpu acceleration, may 2017

\bibitem{kovashka2011actively}
Kovashka, A., Vijayanarasimhan, S., Grauman, K.:
\newblock Actively selecting annotations among objects and attributes.
\newblock In: Proceedings of the IEEE International Conference on Computer
  Vision, IEEE (2011)  1403--1410

\bibitem{deng2013fine}
Deng, J., Krause, J., Fei-Fei, L.:
\newblock Fine-grained crowdsourcing for fine-grained recognition.
\newblock In: Proceedings of the IEEE Conference on Computer Vision and Pattern
  Recognition. (2013)  580--587

\bibitem{linsley2017visual}
Linsley, D., Eberhardt, S., Sharma, T., Gupta, P., Serre, T.:
\newblock What are the visual features underlying human versus machine vision?
\newblock In: Proceedings of the IEEE Conference on Computer Vision and Pattern
  Recognition. (2017)  2706--2714

\bibitem{bucilua2006model}
Buciluǎ, C., Caruana, R., Niculescu-Mizil, A.:
\newblock Model compression.
\newblock In: Proceedings of the 12th ACM SIGKDD international conference on
  Knowledge discovery and data mining, ACM (2006)  535--541

\bibitem{romero2014fitnets}
Romero, A., Ballas, N., Kahou, S.E., Chassang, A., Gatta, C., Bengio, Y.:
\newblock Fitnets: Hints for thin deep nets.
\newblock (2015)

\bibitem{goodfellow2014generative}
Goodfellow, I., Pouget-Abadie, J., Mirza, M., Xu, B., Warde-Farley, D., Ozair,
  S., Courville, A., Bengio, Y.:
\newblock Generative adversarial nets.
\newblock In: Advances in Neural Information Processing Systems. (2014)
  2672--2680

\bibitem{jacobs1991adaptive}
Jacobs, R.A., Jordan, M.I., Nowlan, S.J., Hinton, G.E.:
\newblock Adaptive mixtures of local experts.
\newblock Neural Computation \textbf{3}(1) (1991)  79--87

\bibitem{settles2011theories}
Settles, B.:
\newblock From theories to queries: Active learning in practice.
\newblock In: Active Learning and Experimental Design workshop In conjunction
  with AISTATS 2010. (2011)  1--18

\bibitem{cai2016asking}
Cai, H., Mostofi, Y.:
\newblock Asking for help with the right question by predicting human visual
  performance.
\newblock In: Robotics: Science and Systems. (2016)

\bibitem{cakmak2012designing}
Cakmak, M., Thomaz, A.L.:
\newblock Designing robot learners that ask good questions.
\newblock In: Proceedings of the seventh annual ACM/IEEE international
  conference on Human-Robot Interaction, ACM (2012)  17--24

\bibitem{buck2017ask}
Buck, C., Bulian, J., Ciaramita, M., Gesmundo, A., Houlsby, N., Gajewski, W.,
  Wang, W.:
\newblock Ask the right questions: Active question reformulation with
  reinforcement learning.
\newblock International Conference on Machine Learning (2018)

\end{thebibliography}

\end{document}